\documentclass{svproc}
\usepackage{url}
\usepackage{graphicx}

\begin{document}
\mainmatter              
\title{Efficient Fingerprint Extraction and Matching Using Smartphone Camera}
\titlerunning{Efficient Fingerprint Extraction and Matching Using Smartphone Camera's}  
%
\author{Saksham Gupta \and Sukhad Anand\inst{1}
	\and
Atul Rai\inst{2}
}
\authorrunning{Gupta et al.} 
%
%
\institute{Department of Computer Science, Delhi Technological University, Rohini,Delhi 110042\\
\email{sakshamgupta\_2k14@dtu.ac.in , sukhad.anand@gmail.com}
\and
Staqu Technologies,Gurgoan, Haryana 11004515\\
\email{atual.rai@staqu.com}}

\maketitle              

\begin{abstract}
In the previous decade, there has been a considerable rise in the usage of smartphones.Due to exorbitant advancement in technology, computational speed and quality of image capturing has increased considerably.
With an increase in the need for remote fingerprint verification, smartphones can be used as a powerful alternative for fingerprint authentication instead of conventional optical sensors.
In this research, we propose a technique to capture finger-images from the smartphones and pre-process them in such a way that it can be easily matched with the optical sensor images.Effective finger-image capturing, image enhancement, fingerprint pattern extraction, core point detection and image alignment techniques have been discussed.
The proposed approach has been validated on FVC 2004 DB1 \& DB2 dataset and the results show the efficacy of the methodology proposed. The method can be deployed for real-time commercial usage.
\keywords{Finger-image, Gabor Filter, Core point, Histogram, Remote Verification, FVC}
\end{abstract}
\section{Introduction}
Over the past decade, fingerprint verification has been used irresistibly in various domains including criminal identification, legal matters and user authentication.With  advancement in automated fingerprint identification systems, accompanied by an increasing need for reliable authentication has resulted in widespread deployment of fingerprint authentication in broad applications such as background checks of employees, access to secure facilities, user access authentication in smartphones and remote verification.
\par Remote verification includes verifying the authenticity of a user's located at  access regions. Identity proofing can be  used for diverse applications including secure fund transfer,medical-identity proofing and ATM security.The need for remote verification has been immensely growing especially in developing countries like India which need to verify users located at remote regions for government-related services. One prime example is remote e-KYC(Know your customer Service) started by Indian Government. Generally, these services require  fingerprint authentication which is generally done with optical sensors, difficult to be made accessible in rural areas. With the dawn of the age of smartphones, the overhead of providing optical sensors at outlying regions can be considerably reduced with the help of smartphones.
\par We propose an approach to use smartphone camera to extract fingerprints of a user which can be further used for authentication by matching with images available in any database consisting of optical sensor images.

\section{Related Work}
Substantial research has been done in the field of finger-photo recognition.Some of the considerable work include Derawi et al. \cite{c1} who proposed finger-image recognition with two different smartphones.Neurotechnology, Verifinger 6.0 Extended
SDK commercial minutia extractor has been used for the feature extraction.The verification results of the Neurotechnology
algorithm on the processed images are used for comparison.Stein et al.\cite{c2} proposed a technique for authentication of users on smartphones using finger-image authentication. The proposed method provides algorithms for the capture process to ensure reliable finger-image recognition. Various preprocessing techniques have been described to enhance the recognition rates. In 2015 Anush et. al.\cite{c3} proposed Scatnet based feature representation\cite{c4} and matching algorithm for finger-image images.
\par The above approaches fail to produce reliable results which could be used for remote verification. None of the above-discussed approaches have dealt matching between finger-image captured using a smartphone camera and optical sensor images.
\par \cite{c21} gives a detailed description of various techniques used for Fingerprint Recognition.The paper discusses major techniques used in past for feature extraction. These techniques include Minutiae Based Technique, Pattern Matching or Ridge Feature Based Techniques, Correlation Based Technique and image-based technique.The paper also describes various enhancement techniques used in the past.
\section{Approach}
\begin{figure}
	\includegraphics[width=\linewidth,height=0.5\linewidth]{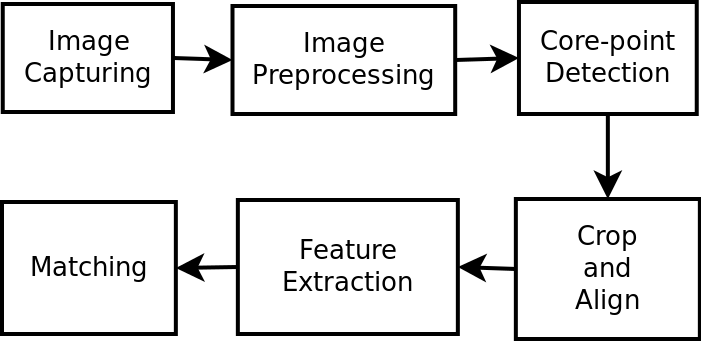}
	\caption{Overview of the proposed methodology}
	\label{overview}
\end{figure}

As shown in Fig.\ref{overview} the proposed fingerprint verification algorithm involves effective image capturing to get a clear and enhanced image.After a clear image has been captured the image has to be preprocessed which involves histogram equalization and removing superfluous noise in the image.After the finger-image has been preprocessed, the fingerprint pattern consisting of frictional ridges has been  extracted so that the finger-image can be further matched with optical sensor images.
After the fingerprint pattern has been extracted, the core point of the finger has to be located so that the finger-image can be properly cropped and aligned with the optical sensor images available.Finally, the image is matched with the database of optical sensor images available.

\subsection{Image Capturing}
\begin{figure}[h!]
	\centering
	\includegraphics[width=0.5\linewidth]{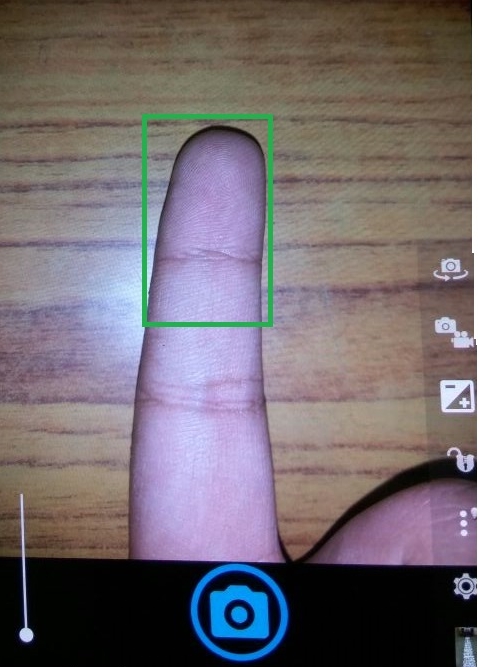}\hfill
	\caption{Custom UI built upon Open Camera Application}
	\label{gui}
\end{figure}

A custom UI application  was developed over Open camera application\cite{c19} for effective image capturing.The UI developed is shown in the Fig.\ref{gui}. The application ensures quality image capturing. Features of the UI include a custom ROI to constrain the location of the fingerprint in the image which eliminates the need for background-separation and finger-image alignment with optical sensor images.
\par Multiple Images are captured in quick succesion by varying exposure during each capture.Finally, an HDR image is extracted which is processed further.
\subsubsection{Region of Intereset(ROI)}
It is a rectangular bounding box which constrains the user to capture finger-image in a particular orientation resulting in desired image alignment.
\subsubsection{High Dynamic Range(HDR)}

A well-known problem in images is due to varied lighting and to ensure that every detail, be it well lit or be it in shadow is captured by the camera we use HDR technique. HDR ensure finger-image quality and capture every minuscule detail. 
We use a five-bracket system in which we vary the exposure of the camera and capture the image at each of these exposures. These five photos are then combined to form a single high detail image.

\subsection{Finger-image Pre-processing}
For precise and unerring authentication using fingerprint matching, the minutiae have to be extracted with an extraordinary amount of precision.For voracious minutiae extraction, the noise from the image has to be completely eliminated i.e the image has to be enhanced.
CLAHE(Contrast limited adaptive histogram equalization)  is superior in terms of accuracy as compared to adaptive histogram equalization because it overcomes its problem of superfluous noise amplification which would have produced infelicitous results. ClAHE captures and improves the local edge information which is essential to the process of fingerprint extraction.
\par
The image is converted from RGB format to a grayscale image and  is partitioned into small non-overlapping regions of equal size called tiles. Histogram equalization is applied to each tile separately. Histogram equalization improves the contrast of the image by stretching out the intensity range at points where pixels have been clustered.
\par
The contrast enhancement can be restricted by restricting the slope of cumulative distribution graph which depends on the slope of the transformation function.This eventually reduces the value of histogram at that pixel value. CLAHE achieves this by clipping the histogram at predefined value which limits the slope of cumulative distribution 
\par
\begin{figure}[h!]
	\centering
	\includegraphics[width=0.48\linewidth,height=.5\linewidth]{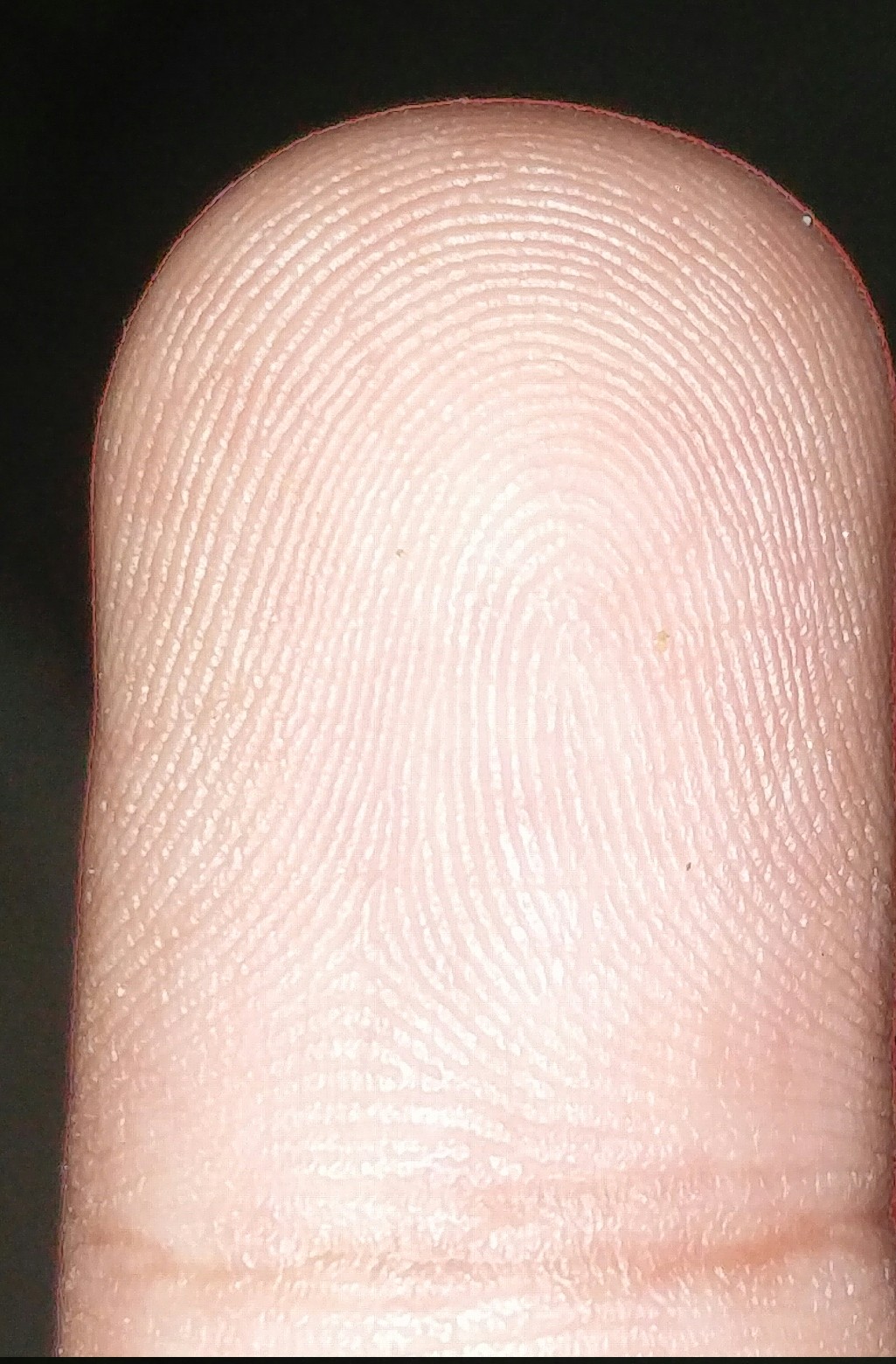}\hfill
	\includegraphics[width=0.48\linewidth,height=.5\linewidth]{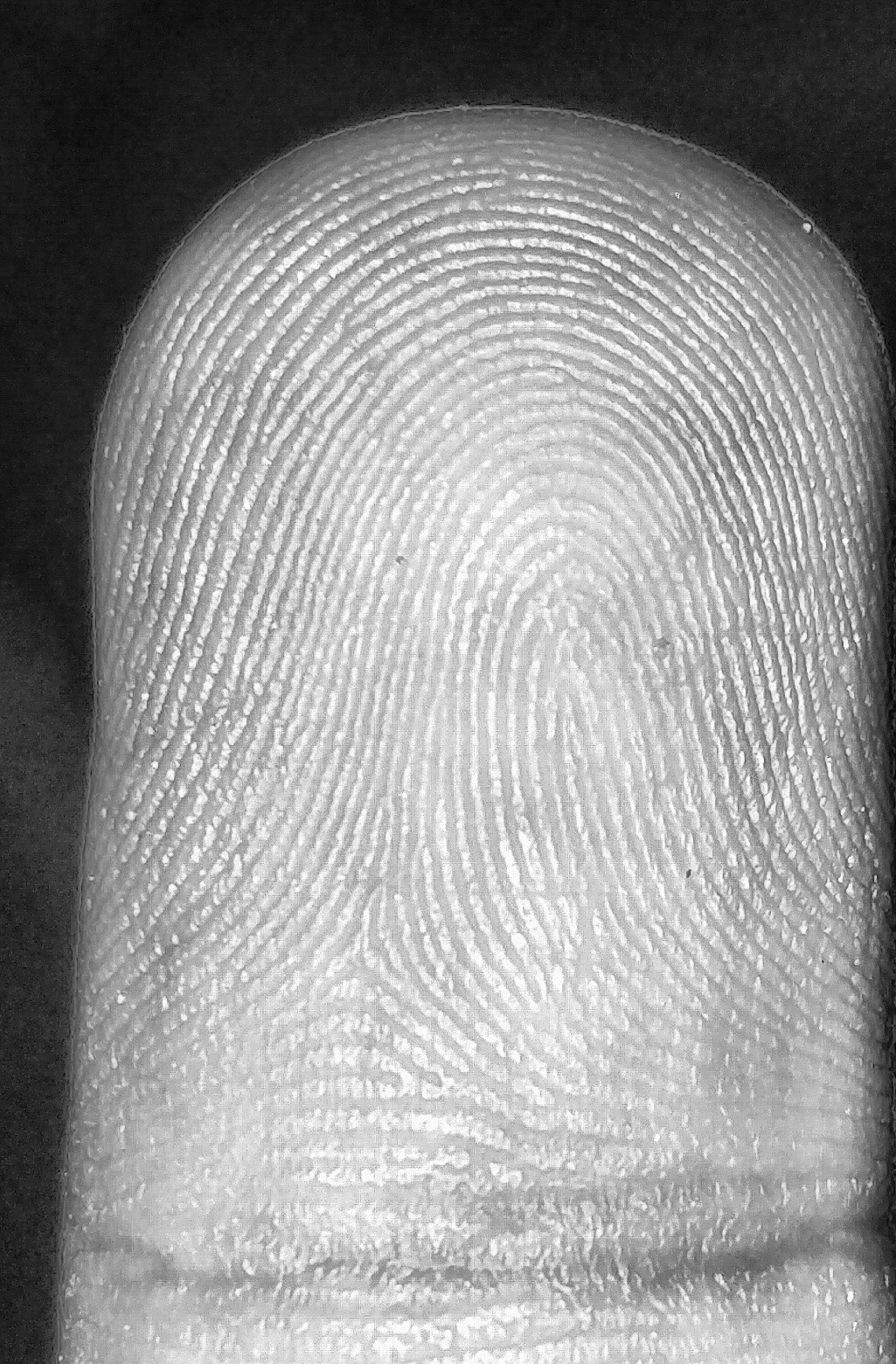}
	\caption{(a)Original Image captured by Custom UI (b)Image after application of pre-processing}
	\label{clahe}
\end{figure}

Superfluous noise in the resultant image is minimized by setting up a clip limit.A clip limit reduces the maximum height of a histogram equalization graph of pixel vs intensity which in turn reduces the amplified noise.After histogram equalization the adjacent tiles are joined together using bilinear interpolation  to completely purge the artificially induced boundaries.
The results obtained after applying CLAHE to the image have been shown in Fig.\ref{clahe}.
\begin{figure}[h!]
	\centering
	\includegraphics[width=0.48\linewidth]{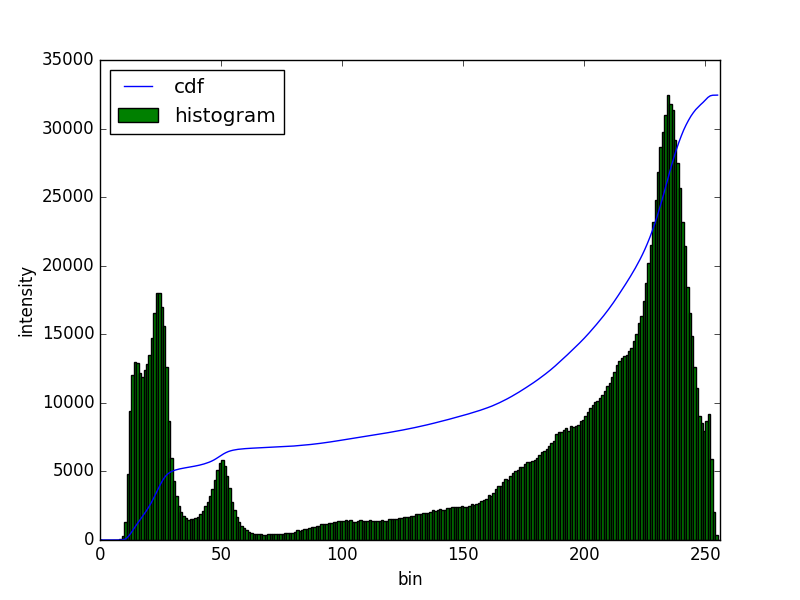}\hfill
	\includegraphics[width=0.48\linewidth]{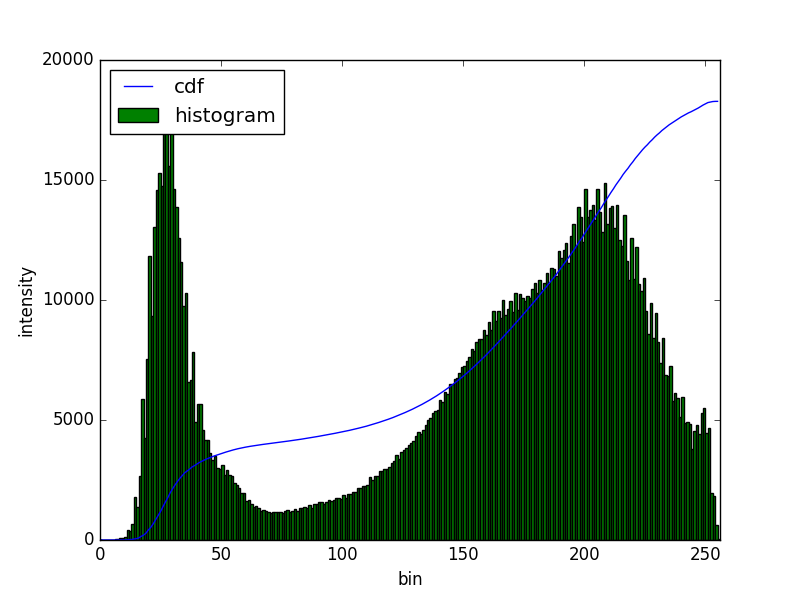}
	\caption{(a)Histogram of Captured image (b) Histogram of Captured Image after application of CLAHE.}
	\label{clahegraph}
\end{figure}
\subsection{Finger-image Pattern Extraction}
To facilitate feature extraction and matching we need to transform our finger-image. Feature Extraction only works well when the image has clear demarcation of ridge-valley structure. To highlight this ridge-valley structure in the finger-image multiple convolutions of 2D Gabor filters varied over range of theta values are used. A detailed explanation of each process is described below:-
\subsubsection{Gabor Filter}
A 2D Gabor filter is defined by sinusoidal plane wave multiplied by a Gaussian function. Discrete domain Gabor filter is given in Eq.\ref{e1},\ref{e2}.
\begin{equation}
G_{c}[i,j] = Be^{-\frac{(i^2 + j^2)}{2\varphi}}\cos(2\pi f(i\cos\Omega+j\sin\Omega))
\label{e1}
\end{equation}
\begin{equation}
G_{s}[i,j] = Ce^{-\frac{(i^2 + j^2)}{2\varphi}}\sin(2\pi f(i\cos\Omega+j\sin\Omega))
\label{e2}
\end{equation}
where $B$ and $C$ are normalising factors, $f$ is frequency in the texture and $\Omega$ gives texture oriented in particular direction, $\varphi$ is the size of the image.Above equations can be derived from 1D Gabor Kernel’s given by
\par $Real$
\begin{equation}
g(a,b;\phi,\Omega,\psi,\varphi,\gamma)=exp(-\frac{a'^2 + \gamma^2 b'^2}{2\varphi^2})\cos(2\pi\frac{a'}{\phi}+\psi)
\end{equation}
\par $Imaginary$
\begin{equation}
g(a,b;\phi,\Omega,\psi,\varphi,\gamma)=exp(-\frac{a'^2 + \gamma^2 b'^2}{2\varphi^2})\sin(2\pi\frac{a'}{\phi}+\psi)
\end{equation}
$where$
\begin{equation}
a' = a\cos\Omega + b\sin\Omega
\end{equation}
$and$
\begin{equation}
b'=-a\sin\Omega + b\cos\Omega
\end{equation}
where $\phi$ is the wavelength of the sinusoidal factor, $\psi$ is the phase offset,$\varphi$is the standard deviation of the Gaussian envelope and $\gamma$ represents the spatial aspect ratio and specifies the ellipticity of the support of the Gabor function.
\subsubsection{Variation in Kernel's}
Multiple kernels are formed by altering the value of theta in Eq.\ref{e1},\ref{e2} above over a range(generally from 0 to pie in steps of pi/16).
\begin{figure}[h!]
	\centering
	\includegraphics[width=0.95\linewidth]{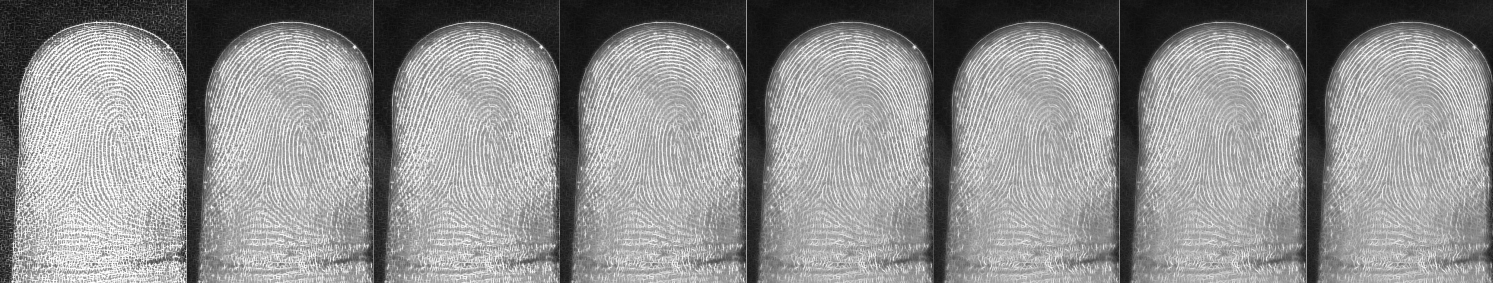}
	\caption{Gabor Convolutions for different Kernel size,left to right:k=15,18,19,20,24,28,30,35}
	\label{kernelsize}
	
\end{figure}
\subsubsection{Convolutions}
Convolution is performed using the kernels generated.Eq.\ref{e3} describe the convolution model used
\begin{equation}
H(x,y) = \sum_{i=0}^{M_i-1}\sum_{j=0}^{M_j-1}I(x+i-a_i, y+j-a_j)K(i,j)
\label{e3}
\end{equation}
Here $I$ is input image $K$ is kernel, $a_i$ and $a_j$ is the value of anchor point in the kernel and $M_i$ and $M_j$ gives kernel size.

\subsection{Core-point Detection}
A crucial determinant in fast fingerprint authentication or verification is the relative alignment of the two fingerprints being matched.  A precise alignment reduces computation by preventing the comparison of each minutiae features set.Additionally the cropping of finger-image is centered around the core-point.
\par In the proposed  work, the latest technique for core point detection proposed by  G. A. Bhagat et al. \cite{c8} has been used.
As the fingerprint consists of frictional ridges.The proposed approach uses ridge orientation map for localizing the core point.
\par The accuracy has been amplified by using an adaptive smoothing technique as described in \cite{c14}. The smoothed orientation map is obtained by 
\begin{equation}
\theta _s(a,b) =\frac{1}{2}\arctan(\frac{\sum_{(m,n)\epsilon\Omega(s)}\sin(2\theta_o(m,n))}{\sum_{(m,n)\epsilon\Omega(s)}\cos(2\theta_o(m,n))})
\end{equation}
where $\theta_s(a,b)$ is the smoothed orientation of the block $(i, j)$,$\Omega(s)$ is the surrounding neighborhood of the block and $s$ is the consistency level and $\theta_o(k,l)$ provides the initial orientation i.e the angle made by ridges passing through neighbourhood centred at (m,n). The computation time has been minimized by dividing the image into smaller blocks of size $r*r$ where $r$ is slightly larger than the width of the frictional ridges.
\par A binary set aligned in square shape is presented which scans the segmented smoothed orientation map. The core point is finally detected using the method proposed by\cite{c8}.
\begin{figure}[h!]
	\centering
	\includegraphics[width=0.5\linewidth]{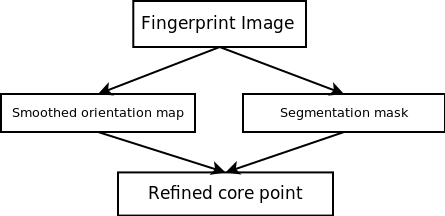}
	\caption{Oveview of Core-point Detection Algorithm}
	\label{corealgopoint}
	
\end{figure}
\subsection{Finger-image Alignment}
After the core point of the fingerprint has been determined, the image has to be properly aligned with the images of the optical sensor images for accurate matching.
\par The image also has to be cropped about the core point. Cropping is necessary because if complete image is given for matching it would result in comparison of large number of non-useful minutiae sets which would ultimately result in increased overhead time and will also result in improper matching.
\par The images obtained from the optical sensor and the finger-image have a phase difference of $180^{0}$.Hence the images captured using smartphones have to be flipped so that the images can be precisely matched.

\subsection{Feature Extraction and Matching}
For the purpose of feature extraction and matching, We used open source application SouceAFIS\cite{c20}. The software is based on MINDTCT template extractor\cite{c15} and Bozorth3 \cite{c16} matching technique. SourceAFIS allow modification in FAR(False Accept Rate) and FRR(False Rejection Rate) so that ERR(Equal Error Rate) can be computed.

\section{Results}
As our approach is based on heterogeneous matching of fingerprints i.e. matching between images prints acquired from the optical sensor and smartphone camera we used the standard optical sensor database FVC 2004(DB1,DB2)\cite{c18} . Finger-images were captured using a smartphone having a 16 MP camera and resolutio of 1920x1080.
\par A custom android application was developed for accurate image acquisition. The application was built upon Open-Camera\cite{c19}.Fig\ref{gui} depicts the application.The bounding box represents the ROI required to enforce alignment.To ensure image quality 5 frames are captured with different exposures in the burst mode and finally an HDR image is generated.
\par
Experiments were performed by transfiguring various parameters, which have been described below:
\subsubsection{Gabor kernel size} 
\begin{figure}[h!]
	\centering
	\includegraphics[width=0.75\linewidth,height=0.5\linewidth]{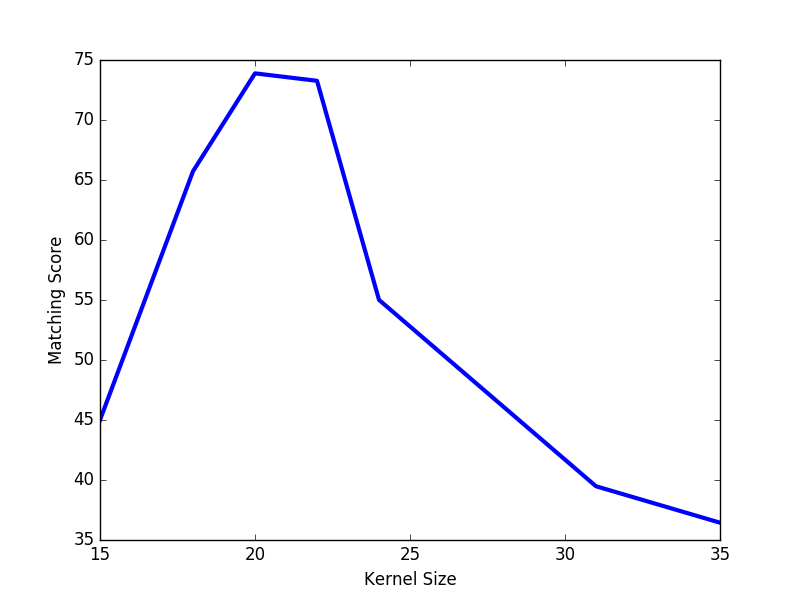}
	\caption{Matching Score for different kernel sizes of 2D Gabour filters}
	\label{kernelscore}
	
\end{figure}

As described above in section\_  multiple Gabor filter convolutions are applied to highlight the ridge-valley structure of the print. Gabor filter is applied to the images in the form of a 2-D convolution kernel. Experiments were made upon different Gabor kernel sizes to ensure maximum matching accuracy. The curve shown in Fig \ref{kernelscore} depicts the relationship between various Kernel sizes and the matching scores obtained using SourceAFIS matcher.

\par
From the curve, we conclude that the matching score achieve its maxima for a kernel size of 21 with a score of 74.23. Thus a Gabor filter of square kernel size of 21 supplies best results for our experiments. Further results are computed using this kernel size only.

\subsubsection{Effective area of interest around the core point}
\begin{figure}[h!]
	\centering
	\includegraphics[width=0.48\linewidth,height=0.48\linewidth]{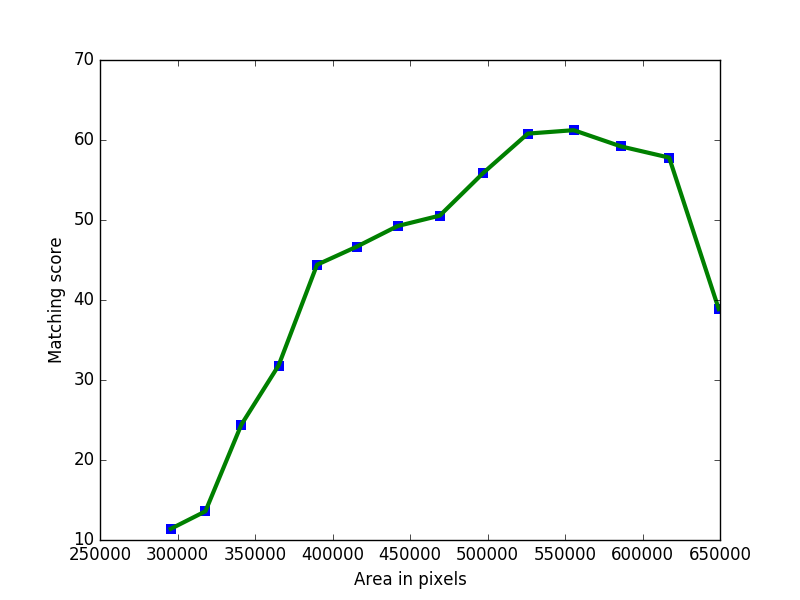}
	\includegraphics[width=0.48\linewidth,height=0.48\linewidth]{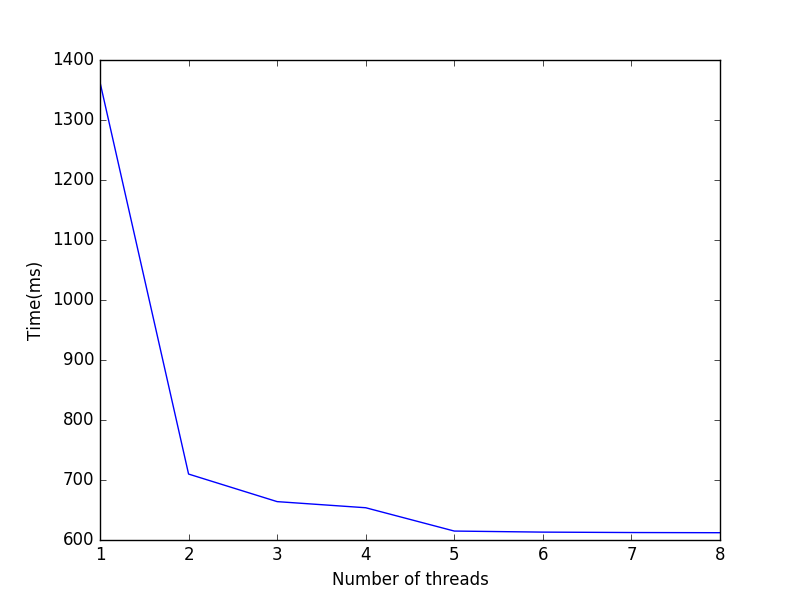}
	\caption{(a)Matching Score(SourceAFIS) varied over area of cropping(b)Processing Time required to find Gabor convolutions over multiple threads.}
	\label{speedup}
	
\end{figure}

After core point detection, different dimensions for cropping area around the core point were considered to find its dependence on matching score.Fig \ref{speedup}(a) represents the desired dependency relationship.

From the curve, we can ascertain that the optimal dimensions of the effective region of interest to achieve maximum accuracy are 800x700 pixels around the core point detected. 
\subsubsection{Threading}
Our approach consists of convolutions of multiple Gabor filters which is a computationally expensive activity. For a multi-processor system, applying threading can significantly speed up the convolution process. Fig \ref{speedup}(b) depicts the time of execution for different number of threads. From the graph it is evident that computation time decreases drastically as the number of threads are increased and saturates around a thread count of 5.

\subsubsection{Accuracy}
\begin{figure}[h!]
	\centering
	\includegraphics[width=0.48\linewidth,height=0.48\linewidth]{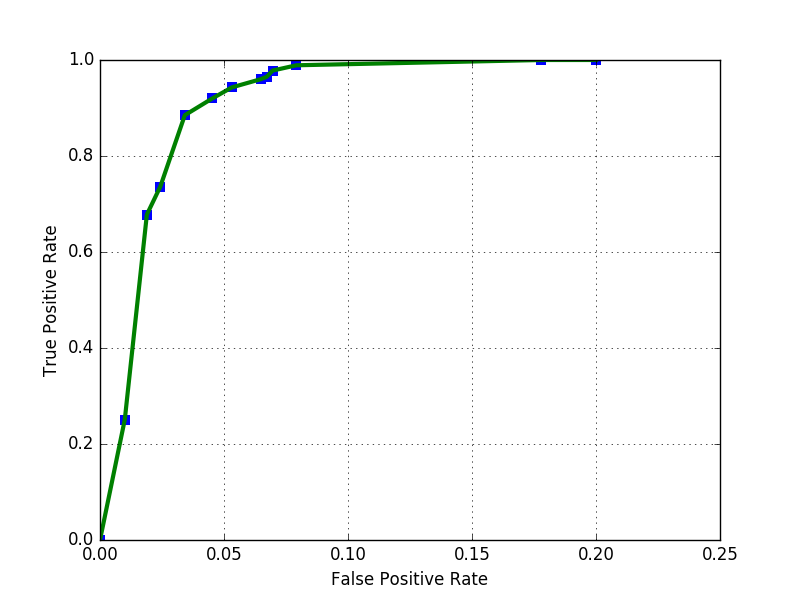}
	\caption{ROC curve obtained by varying threshold values of the matcher}
	\label{roc}
	
\end{figure}

The ROC curve of the experiments performed with different thresholds of the matcher is shown in Fig.\ref{roc}.As the value of the threshold is increased,  both true positive rate and the false positive rate decrease and vice versa. ERR for the proposed technique on the FVC 2004 database lies between the range of 6\% - 15\%, thus making the approach feasible for commercial usage in real-time.

\section{Conclusion and Future Work}
By this study, we conclude that fingerprint authentication can be performed using Smartphone camera’s. We were able to match the finger-images obtained using the camera with the fingerprints obtained using the optical sensor. The EER was 6\% - 15\% which is well within the acceptable limits. Thus this technique can be efficiently used by people not having easy access to fingerprint sensors.
The technique can  be greatly improved by adjusting the Gabor filter parameters optimal to the setting of finger-image capturing. Further better minutiae extractor and matcher can be used to accurately match the prints. As the technology evolves more powerful smartphone camera’s shall be able to capture higher resolution images which will ultimately lead to better extraction of the ridge-valley structure.

\end{document}